\documentclass[runningheads]{llncs}
\usepackage{graphicx}
\usepackage{comment}
\usepackage{amsmath,amssymb} % define this before the line numbering.
\usepackage{color}
\usepackage{multirow}
\usepackage{booktabs}
\usepackage{subfigure}
\usepackage{wrapfig}
\usepackage{parskip}
\usepackage[linesnumbered,boxed,ruled,commentsnumbered]{algorithm2e}
\usepackage{wrapfig}

\usepackage[amsmath,thmmarks]{ntheorem}

\theoremseparator{.}
\theoremstyle{empty}

\usepackage{floatrow}
\newfloatcommand{capbtabbox}{table}[][\FBwidth]

\begin{document}

% \linenumbers
\pagestyle{headings}
\mainmatter
\def\ECCVSubNumber{2330}  % Insert your submission number here

% \title{Semi-supervised crowd counting via semi-supervised learning on auxiliary prediction tasks} % Replace with your title
\title{Semi-Supervised Crowd Counting via Self-Training on Surrogate Tasks}

\begin{comment}
\titlerunning{ECCV-20 submission ID \ECCVSubNumber} 
\authorrunning{ECCV-20 submission ID \ECCVSubNumber} 
\author{Anonymous ECCV submission}
\institute{Paper ID \ECCVSubNumber}
\end{comment}
%******************

%\begin{comment}
\titlerunning{Semi-Supervised Crowd Counting via Self-Training on Surrogate Tasks}
\author{Yan Liu\inst{1} \and
Lingqiao Liu\inst{2} \thanks{The first two authors have equal contribution.}\and
Peng Wang\inst{3} \and
Pingping Zhang\inst{4} \and
Yinjie Lei\inst{1} \thanks{The corresponding author: Yinjie Lei (Email: yinijie@scu.edu.cn).} }

%\renewcommand{\thefootnote}{\fnsymbol{footnote}}
%\footnotetext[1]{You can add acknowledgements here.}

%
\authorrunning{Y. Liu et al.}
\institute{College of Electronics and Information Engieering, Sichuan University\\
\email{yanliu27@stu.scu.edu.cn, yinjie@scu.edu.cn}
\and
School of Computer Science, The University of Adelaide\\
\email{lingqiao.liu@adelaide.edu.au}
\and
School of Computing and Information Technology, University of Wollongong\\
\email{pengw@uow.edu.au}
\and
School of Artificial Intelligence, Dalian University of Technology\\
\email{jssxzhpp@mail.dlut.edu.cn}}

%\end{comment}
%******************
\maketitle

\begin{abstract}
Most existing crowd counting systems rely on the availability of the object location annotation which can be expensive to obtain. To reduce the annotation cost, one attractive solution is to leverage a large number of unlabeled images to build a crowd counting model in semi-supervised fashion. This paper tackles the semi-supervised crowd counting problem from the perspective of feature learning. Our key idea is to leverage the unlabeled images to train a generic feature extractor rather than the entire network of a crowd counter. The rationale of this design is that learning the feature extractor can be more reliable and robust towards the inevitable noisy supervision generated from the unlabeled data. Also, on top of a good feature extractor, it is possible to build a density map regressor with much fewer density map annotations. Specifically, we proposed a novel semi-supervised crowd counting method which is built upon two innovative components: (1) a set of inter-related binary segmentation tasks are derived from the original density map regression task as the surrogate prediction target; (2) the surrogate target predictors are learned from both labeled and unlabeled data by utilizing a proposed self-training scheme which fully exploits the underlying constraints of these binary segmentation tasks. Through experiments, we show that the proposed method is superior over the existing semi-supervised crowd counting method and other representative baselines.

\end{abstract}

\section{Introduction}

Crowd counting is to estimate the number of people or objects from images or videos. Most existing methods formulate it as a density map regression problem \cite{lempitsky2010learning,zhang2015cross,zhang2016single,sam2017switching,li2018csrnet:}, and solve it by using the pixel-to-pixel prediction networks \cite{kang2014fully,peng2017large,long2015fully}. Once the density map is estimated, the total object count can be trivially calculated. To train such a density map regression model, most existing crowd counting methods rely on a substantial amount of labeled images with the object location annotation, e.g., marking a dot at the center of corresponding persons. The annotation process can be labor-intensive and time-consuming. For example, to annotate the ShanghaiTech \cite{zhang2016single} dataset, 330,165 dots must be placed on corresponding persons carefully. 

To reduce the annotation cost, an attractive solution is to learn the crowd counter in a semi-supervised setting which assumes availability of a small amount of labeled images and a large amount of unlabeled images. This is a realistic assumption since unlabeled images are much easier or effortlessly to obtain than labeled images. Then the research problem is how to leverage the unlabeled image to help train the crowd counter for achieving a reasonable performance. 

To solve this problem, we propose a novel semi-supervised learning algorithm to obtain a crowd counting model. One key of our model is to use the unlabeled data to learn a generic feature extractor of the crowd counter instead of the entire network as most traditional methods do. The underlying motivations are threefold: (1) It is challenging to construct a robust semi-supervised learning loss term from unlabeled data for regression output. In contrast, learning a feature extractor is more robust and reliable towards the inevitable noisy supervision generated from unlabeled data; (2) the feature extractor often plays a critical role in a prediction model. If we have a good feature extractor, it is possible to learn a density map regressor, i.e., crowd counter, require much less ground-truth density map annotations; (3) there are a range of methods for learning feature extractor, and features can be even learned from other tasks rather than density map regression (i.e., surrogate tasks in this paper). 

Inspired by those motivations, we propose to learn the feature extractor through a set of surrogate tasks: predicting whether the density of a pixel is above multiple predefined thresholds. Essentially, those surrogate tasks are binary segmentation tasks and we build multiple segmentation predictors for each of them. Since those tasks are derived from the density map regression, we expect that through training with these surrogate tasks the network can learn good features to benefit the density map estimation. For labeled images, we have ground-truth segmentation derived from the ground-truth density map. For unlabeled images, the ground-truth segmentation are not available. However, the unlabeled images can still be leveraged through a semi-supervised segmentation algorithm. Also, we notice that the correct predictions for the surrogate tasks should hold certain inter-relationship, e.g., if the density of a pixel is predicted to be higher than a high threshold, it should also be predicted higher than a low threshold. Such inter-relationships could serve as additional cues for jointly training segmentation predictors under the semi-supervised learning setting. Inspired by that, we developed a novel self-training algorithm to incorporate these inter-relationships to generate reliable pseudo-labels for semi-supervised learning. By conducting extensive experiments, we demonstrate the superior performance of the proposed method. To sum up, our main contributions are:

\begin{itemize}
\item We approach the problem of semi-supervised crowd counting from a novel perspective of feature learning. By introducing the surrogate tasks, we cast the original problem into a set of semi-supervised segmentation problem.
\item We develop a novel self-training method which fully takes advantage of the inter-relationship between multiple binary segmentation tasks. 

\end{itemize}

\section{Related Works}

\noindent \textbf{Traditional Crowd Counting Methods} include detection-based and regression-based methods. The detection-based methods use head or body detectors to obtain the total count in an image \cite{subburaman2012counting,dalal2005histograms,viola2004robust}. However, in extremely congested scenes with occlusions, detection-based methods can not produce satisfying predictions. 
 
Regression-based methods are proposed \cite{chen2012feature,chan2009bayesian} to tackle challenges in overcrowded scenes. In regression-based methods, feature extraction mechanisms \cite{Idrees_2013_CVPR,fiaschi2012learning} such as Fourier Analysis and Random Forest regression are widely used. However, traditional methods can not predict total counts accurately as they overlook the spatial distribution information of crowds. 

\noindent \textbf{CNN-based Crowd Counting Methods} learn a mapping function from the semantic features to density map instead of total count \cite{lempitsky2010learning}. Convolutional Neural Network (CNN) shows great potential in computer vision tasks. CNN-based methods are used to predict density maps. Recently, the mainstream idea is to leverage deep neural networks for density regression \cite{zhang2015cross,zhang2016single,sam2017switching,Sindagi2017generate}. These methods construct multi-column structures to tackle scale variations. Then local or global contextual information is obtained for producing density maps. 

Several works \cite{li2018csrnet:,chen2019scale} combine the VGG \cite{simonyan2014very} structure with dilated convolution to assemble the semantic features for density regression. While other works \cite{jiang2019crowd,zhao2019leveraging,Ranjan_2018_ECCV,zhu2019dual} introduce attention mechanisms to handle several challenges, e.g. background noise and various resolutions. Meanwhile works \cite{Liu2018decidenet,lian2019density,jiang2019mask,valloli2019wnet,Idrees_2018_ECCV} leverage the multi-task frameworks, i.e., detection, segmentation or localization, which provide more accurate location information for density regression. Besides, the self-attention mechanism \cite{zhang2019relational,wan2019adaptive} and residual learning mechanism \cite{sindagi2019pushing} are effective in regularizing the training of the feature extractor. Work \cite{xiong2019open} transforms the density value to the density level from close-set to open-set. Further, a Bayesian-based loss function \cite{ma2019bayesian} is proposed for density estimation. These above CNN-based methods require a large number of labeled images to train the crowd counter. However, annotating the crowd counting dataset is a time-consuming and labor-intensive work.

\noindent \textbf{Semi-/Weakly/Un-Supervised Crowd Counting Methods} attempt to reduce the annotation burden by using semi-/weakly/un-supervised settings. In the semi-supervised setting, work \cite{liu2018leverageing} collects large unlabeled images as extra training data and constructs a rank loss based on the estimated density maps. In the weakly-supervised setting, work in \cite{lei2020weakly} establishes a scheme of multiple auxiliary tasks training on the large amount of count-level annotations. Also, work in \cite{von2016gaussian} leverages the total count as a weak supervision signal for density estimation. Besides, an auto-encoder structure \cite{sam2019almost} is proposed for crowd counting in an almost unsupervised setting. Another method for reducing the annotation burden is to use synthetic images \cite{wang2019learning}. For example, the GAN-based \cite{gao2019featureaware} and domain adaption based \cite{gao2019domainadaptive} frameworks combine the synthetic images and realistic images to train the crowd counter. These methods are effective in reducing the annotation burden. However, they can not obtain satisfying crowd counting performance because the inevitable noisy supervision may mislead the density regressor.

\vspace{-30pt} 
 \begin{figure}
\centering \subfigure[Traditional semi-supervised methods.]{ \includegraphics[width=5.8cm]{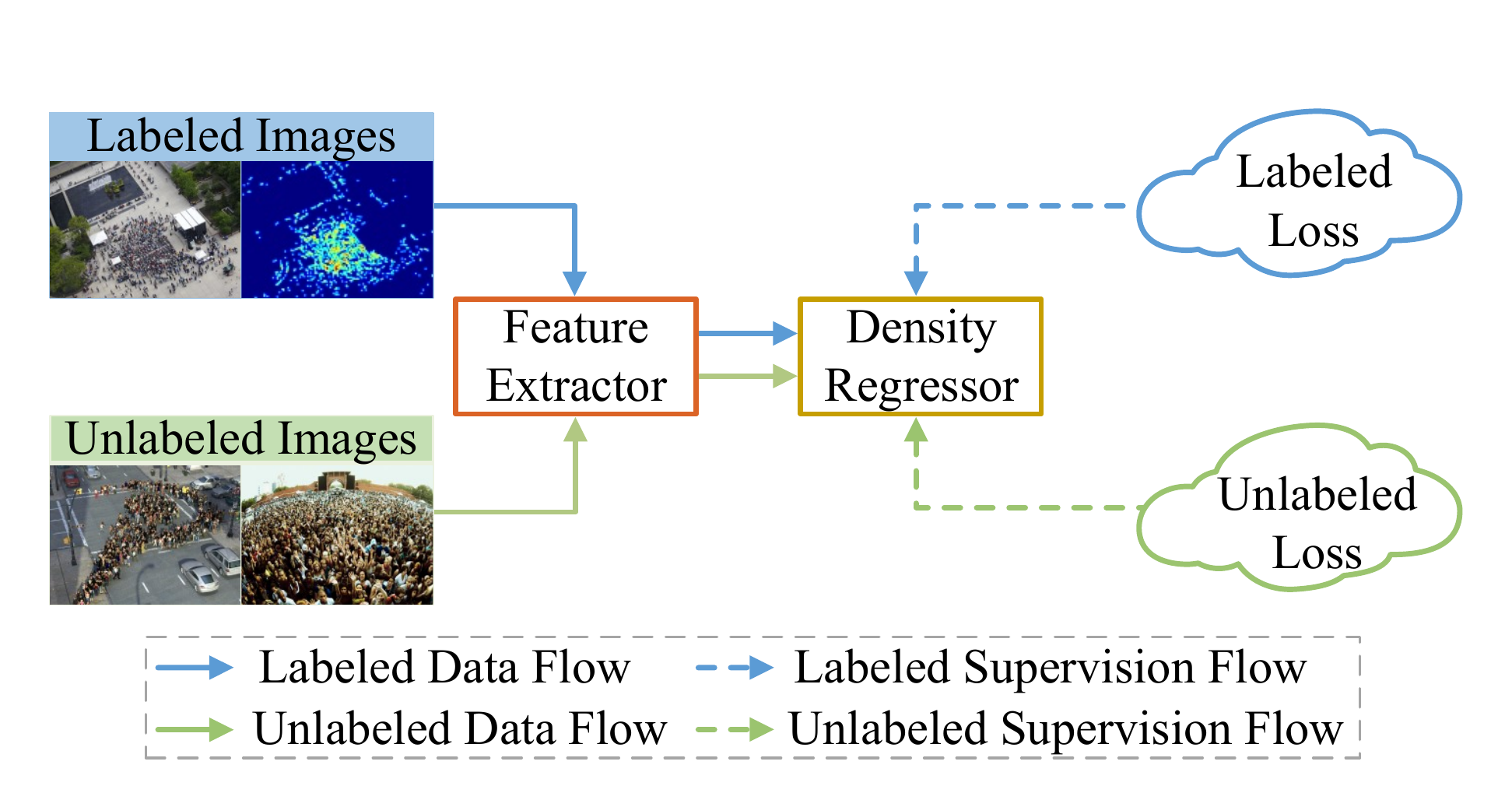}
\label{fig:sub_a} } 
\subfigure[Proposed semi-supervised method.]{
\includegraphics[width=5.8cm]{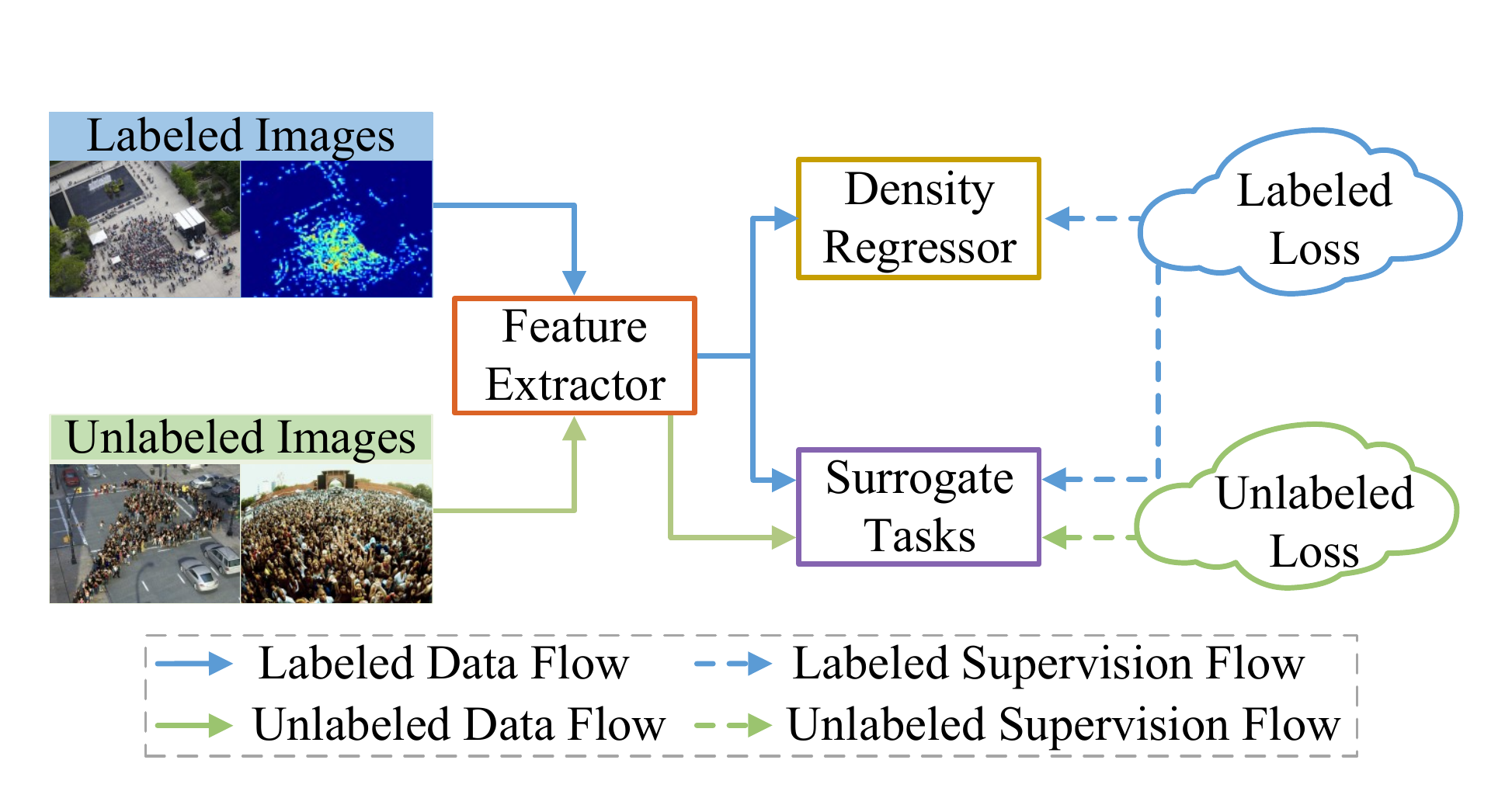} \label{fig:sub_b} }
\caption{(a) Traditional semi-supervised methods use both labeled and unlabeled images to update the feature extractor and density regressor. (b) In the proposed method, the unlabeled images are only used for updating feature extractor.} 
\label{fig:motivation} 
\vspace{0pt} 
\end{figure}

\section{Background: Crowd Counting as Density Estimation\label{sec:background}}

Following the framework ``learning to count" \cite{lempitsky2010learning}, crowd counting can be transformed into a density map regression problem. Once the density map is estimated, the total object count can be simply estimated by its summation, that is, $\hat{N} = \sum_{i,j} \hat{D}(i,j)$, where $\hat{D}(i,j)$ denotes the density value for pixel $(i,j)$. The Mean Square Error (MSE) loss is commonly used in model training, that is,
\begin{equation}
\mathcal{L}_{MSE}=\sum_{(i,j)}|\hat{D}(i,j)-D(i,j)|^{2},\label{eq:MSE-loss}
\end{equation}
where $\hat{D}$ is the estimated density map and $D$ is the ground-truth density map.

\begin{figure}[tbh]
\centering{}\includegraphics[width=12.4cm]{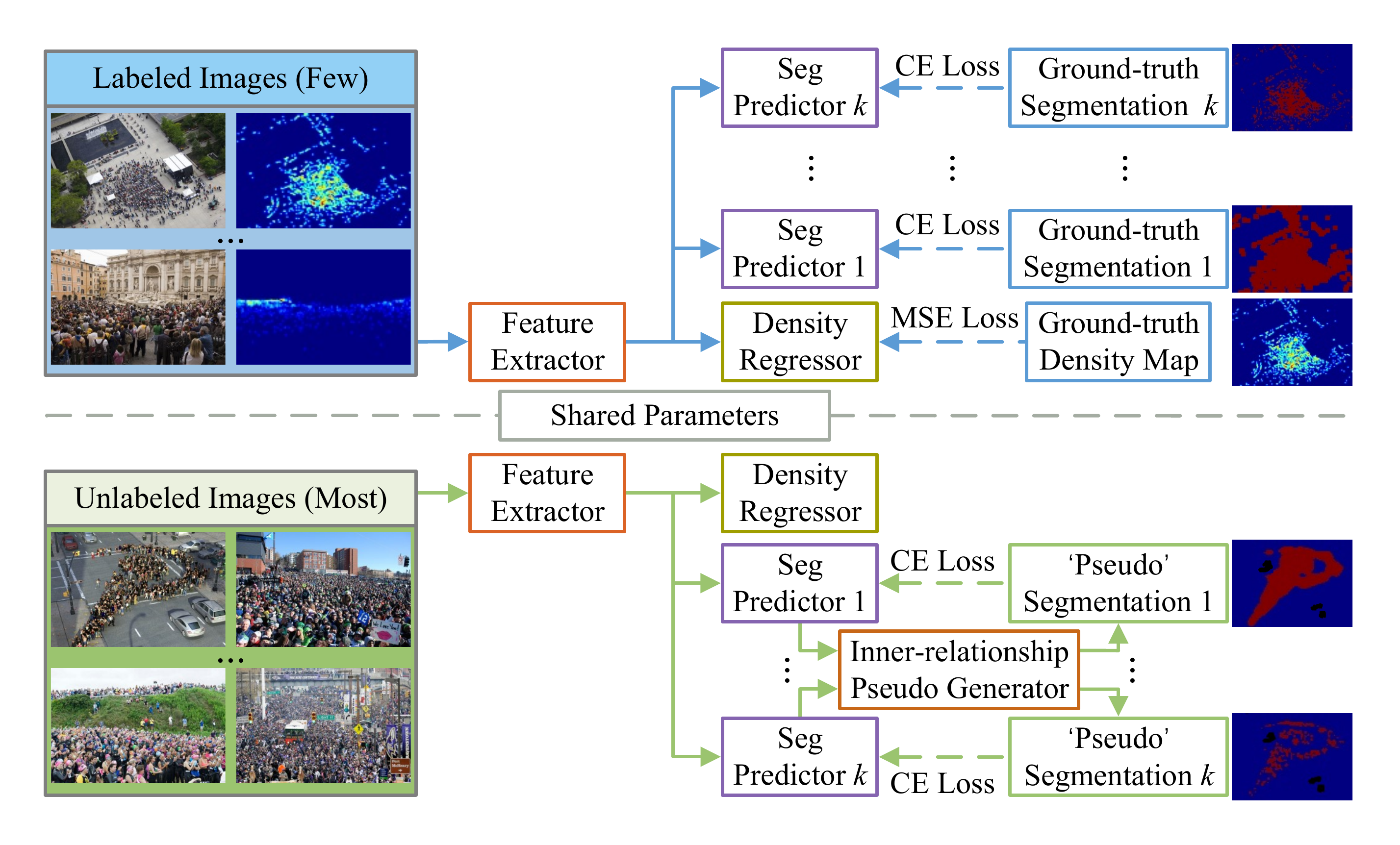}\caption{ The overview of our proposed method. We introduce a set of binary segmentation surrogate tasks. For labeled images, we construct loss terms on both original and surrogate tasks. For unlabeled images, we use the output of segmentation predictor and inter-relationship to generate ``pseudo segmentation", which is shown in Figure. \ref{fig:interrelationship}.\label{fig:overview-of-the-proposed-method}}
\end{figure}

\section{Methodology}
In this paper, we are interested in learning a crowd counter based on the semi-supervised setting. Formally, we assume that we have a set of labeled images $L=\{I^l_i, D_i\}$, where $D_i$ is the ground-truth density map, and a set of unlabeled images $U=\{I^u_i\}$. Our task is to learn a crowd counter by using both the labeled images and unlabeled images. In our setting, the unlabeled set contains much more images than the labeled set for training a crowd counting model.

\subsection{Using Unlabeled Data for Feature Learning}\label{sect:feature_training}
Generally speaking, a network can be divided into two parts, a feature extractor and a task-specific predictor. The former converts the raw images into feature maps while the latter further transforms them to the desired output, e.g., density map, in the context of crowd counting. Most existing semi-supervised learning methods \cite{xie2019unsupervised,liu2018leverageing,tarvainen2017mean,verma2019interpolation} learn those two parts simultaneously and seek to construct a loss term from unlabeled data applied to the entire network. 

In contrast to the existing methods, we propose to learn the feature extractor and the task-specific predictor through different tasks and loss terms. In particular, in our method, the unlabeled data is only used for learning the feature extractor. This design is motivated by three considerations: (1) crowd counting is essentially a semi-supervised regression problem in our setting. Besides, it can be challenging to construct a robust semi-supervised regression loss term from unlabeled data (i.e., as most existing methods do). The noisy supervision generated from the loss term from unlabeled data may contaminate the task-specific predictor and lead to inferior performance. In our method, unlabeled data is only used to train the feature extractor as the noisy supervision will not directly affect the task-specific predictor; (2) feature extractor plays an important role in many fields like unsupervised feature learning \cite{zhang2014saliency,dosovitskiy2014discriminative}, semi-supervised feature learning \cite{yang2013semi,cheng2014semi} and few-shot learning \cite{gidaris2019boosting,li2019finding}. Indeed, with a good feature extractor, it is possible to reduce the need of a large amount of labeled data in training. In the context of crowd counting, this implies that much less ground-truth density map annotations are needed if we can obtain a robust feature extractor via other means; (3) feature extractor can be learned in various ways. In this way, we will have more freedom in designing semi-supervised learning algorithms for feature learning. Specifically, we propose to derive surrogate tasks from the original density map regression problem, and use those tasks for training the feature extractor. The schematic overview of this idea is shown in Figure \ref{fig:motivation} (b). For labeled images, the target of surrogate task can be transformed from ground-truth annotation. For the unlabeled images, the ground-truth annotation becomes unavailable. However, the unlabeled images can still be leveraged to learn the surrogate task predictor and consequently the feature extractor in a semi-supervised learning manner. In the following sections, we first elaborate how to construct the surrogate loss and then describe the semi-supervised learning algorithm developed for the surrogate tasks.

\subsection{Constructing Surrogate Tasks for Feature Learning \label{sec:surrogate-task}}

The surrogate task defined in this paper is to predict whether the density value of a pixel, $D(i,j)$, exceeds a given threshold. In other words, the prediction target of the surrogate task is defined as:

\begin{equation}
M(i,j)=\begin{cases}
1~~~~~&D(i,j)>\epsilon \\0~~~~~&D(i,j)<=\epsilon \end{cases}
,\label{eq:density_to_mask}
\end{equation}
where $(i,j)$ is the pixel coordinate, and $\epsilon$ is the predefined threshold. For labeled data, the ground-truth of $D$ is known and thus $M$ is known. For unlabeled data, no annotation of $D$ is available and thus $M$ is unknown. However, we can still use unlabeled data to construct loss term for indirectly supervising the prediction of $M$. Note that in this way, we essentially recast the original semi-supervised crowd counting problem into a semi-supervised segmentation problem since $M$ only takes binary values.   

In practice, we use multiple thresholds and generate multiple surrogate targets $\{M_k\}$ to consider the pixels with different density levels. To set these thresholds, we rank all non-zero density values from all the labeled images in ascending order and choose the thresholds as the value ranked at $r_k\times N$, where $r_k \in [0,1]~ k=1,..,c$, $N$ is the total number of non-zero values and $c$ indicates the number of surrogate tasks. Meanwhile, we create multiple segmentation predictor branches attached to the feature extractor. These surrogate tasks are parallel to the density map regressor, as shown in Figure \ref{fig:overview-of-the-proposed-method}. 

\subsection{\underline{I}nter-\underline{R}elationship-\underline{A}ware \underline{S}elf-\underline{T}raining (IRAST) for Semi-supervised Training on Surrogate Tasks\label{sec:inner-aware}}

\begin{figure*}[tbh]
\vspace{-20pt}
\includegraphics[width=12.0cm]{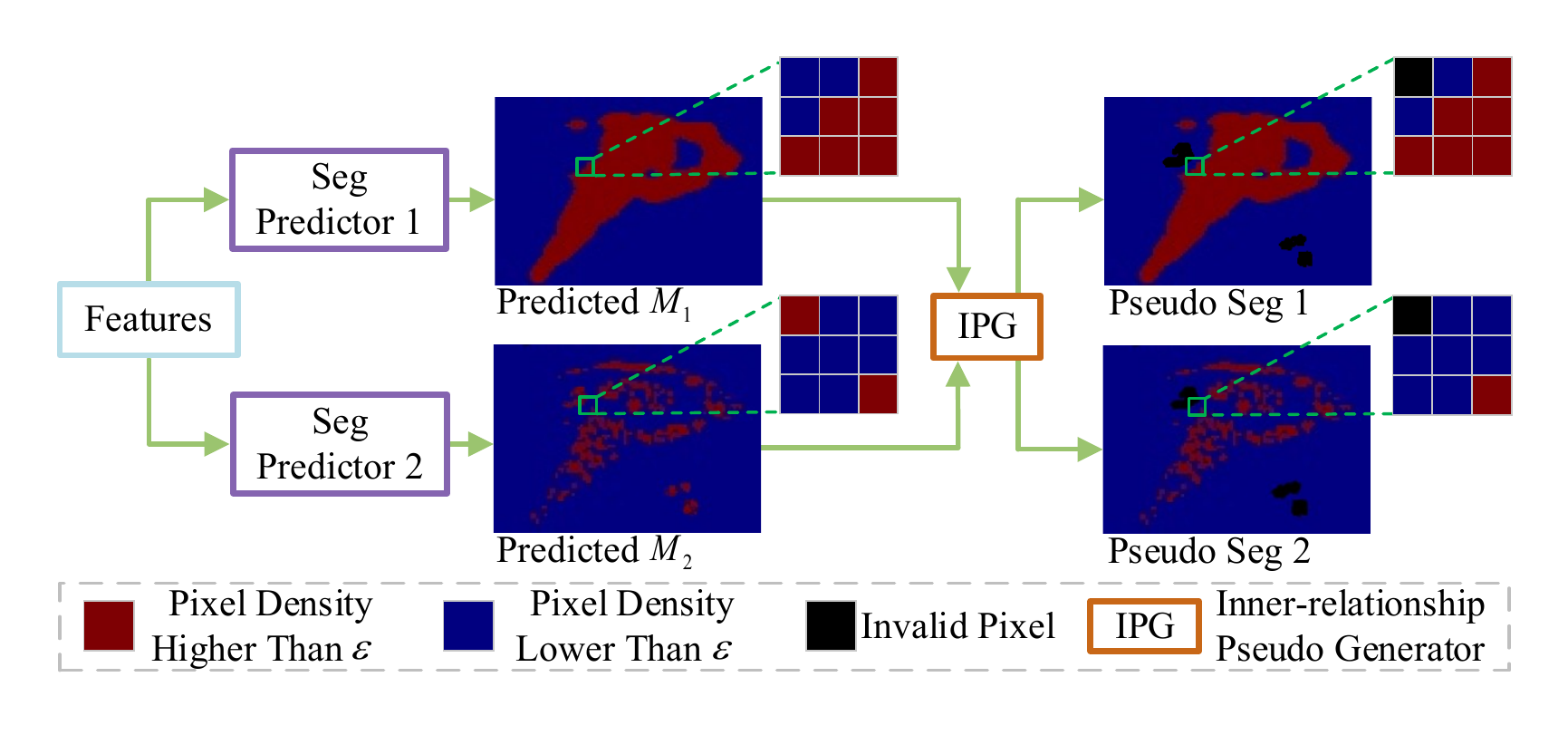}\caption{The illustration of the inter-relationship between two segmentation predictors. We use a lower threshold $\epsilon_{1}$ segmentation predictor to produce $\hat{M}_{1}$, and a higher threshold $\epsilon_{2}$ segmentation predictor to produce $\hat{M}_{2}$. If a specific pixel in $\hat{M}_{1}$ is lower than $\epsilon_{1}$, while in $\hat{M}_{2}$ is higher than $\epsilon_{2}$, we can consider this pixel is invalid. The inter-relationship avoids such incorrect training signal flowing into the feature extractor.  \label{fig:interrelationship}}
\end{figure*}

\IncMargin{1em}
\begin{algorithm}[h] 

\KwIn{ Number of surrogate tasks $c$. Given threshold $t_p$,  
Predicted confidence value (posterior probability)  $P(\hat{M}_{k}=1)~k=1,\cdots,c$; $P(\hat{M}_{k}=0) = 1 - P(\hat{M}_{k}=1)$ }
\KwOut{ A set of pseudo-label set $\{\mathcal{S}_k\}$, one for each $k$:  $\mathcal{S}_k=\{(i,j,s_{ij})\}$, where $s_{ij}$ is the generated pseudo-label for $(i,j)$.}

\For{$k \in [1,c]$}
    {\For{ each location $(i,j)$}
        {\If{$P(\hat{M}_{k}(i,j)=1)>t_p$ and $P(\hat{M}_g(i,j)=1)>t_p~~ \forall g<k$}
            {$\mathcal{S}_k \leftarrow \mathcal{S}_k \cup (i,j,1)$}
        
         {\If{$P(\hat{M}_{k}(i,j)=0)>t_p$ and $P(\hat{M}_h(i,j)=0)>t_p~~ \forall h>k$}
            {$\mathcal{S}_k \leftarrow \mathcal{S}_k \cup (i,j,0)$}}
     
        }}

\caption{Pseudo-label Generation Rule \label{alg:pesudo-generate}}    
\end{algorithm}
\DecMargin{1em}

To leverage the unlabeled data to train the surrogate task predictors and the feature extractor, a semi-supervised learning algorithm is needed. Self-training is one of the most commonly used semi-supervised learning algorithms in segmentation tasks \cite{socher2010connecting,karnyaczki2015sparse}. It recursively generates pseudo-class-label for samples (pixels) with prediction confidence values higher than a given threshold $t_p=0.9$. However, this straightforward solution largely ignores the underlying inter-relationship between multiple surrogate tasks.
Recall that $M$ takes binary values and $M(i,j) = 1$ if the density value of pixel $(i,j)$ is greater than a given threshold. Suppose we have two segmentation results $M_1$ and $M_2$ estimated from two predictors corresponding to two thresholds $\epsilon_1$ and $\epsilon_2$ ($\epsilon_1 < \epsilon_2$), then there will be a conflict if one predictor gives the prediction $\hat{M}_1(i,j) = 0$ while the other gives the prediction $\hat{M}_2(i,j) = 1$. This is because $\hat{M}_1(i,j) = 0$ indicates the density value of the pixel is less than $\epsilon_1$, but $\hat{M}_2(i,j) = 1$ implies the density value of pixel is larger than $\epsilon_2$ and consequently larger than $\epsilon_1$ since $\epsilon_1 < \epsilon_2$.

This inter-relationship could essentially act as an error correcting mechanism to test if the prediction made by surrogate predictors are likely to be accurate. Thus in our method, we incorporate it into the framework of self-training as an additional criterion for pseudo-label generation besides the commonly used thresholding criterion. Formally, we define the following rule for generating a pseudo label at the $k$-th predictor. Without loss of generality, we assume there are $c$ predictors, ranking from $1$ to $c$ according to the descent order of their corresponding thresholds, that is, $\epsilon_a > \epsilon_b$ if $a > b$. The formal rule of generating pseudo-label is shown in Algorithm \ref{alg:pesudo-generate}.The generation of pseudo-labels is online.

In nutshell, a pseudo label is generated in the surrogate binary segmentation task if its prediction confidence value for one class (``1'' or ``0'' in our case) is greater than $t_p$ and its prediction is not conflict with predictions of other predictors. An example of this scheme is illustrated in Figure \ref{fig:interrelationship}.

\noindent \textbf{Discussion:} The proposed method defines $c$ binary segmentation tasks and one may wonder why not directly define a single $c$-way multi-class segmentation task. Then an standard multi-class self-training method can be used. We refer this method as \textbf{Multiple-class Segmentation Self-Training (MSST)}. Comparing with our approach, MSST has the following two disadvantages: (1) it does not have the ``error correction'' mechanism as described in the rule of generating pseudo label. The difference between MSST and IRAST is the standard one-vs-rest multi-class classification formulation and the error correcting output codes formulation \cite{dietterich1994solving}; (2) MSST may be overoptimistic towards the confidence score due to the softmax normalization of logits. Considering a three-way classification scenario for example, it is possible that the confidence for either class is low and the logits for all three classes are negative. But by chance, one class has relatively larger logits, say, $\{-100, -110$ and $-90\}$ for class 1, 2 and 3 respectively. After normalization, the posterior probability for the last class becomes near 1, and will exceed the threshold for generating pseudo labels. In contrast, the proposed IRAST does not have this issue since the confidence score will not be normalized across different classes (quantization level). We also conduct an ablation study in Section \ref{sec:msst-vs-irast} to verify that MSST is inferior to IRAST.

\section{Overall Training Process}
In practice, we use the Stochastic Gradient Descent (SGD) to train the network\footnote{As the unlabeled set contains more images than the labeled set, we oversample labeled images to ensure the similar amount of labeled and unlabeled images occur in a single batch.}. For an labeled image, we construct supervised loss terms based on the density regression task and surrogate tasks, and the training loss is:
\small{{\begin{align}
\mathcal{L}_{L} & =\mathcal{L}_{MSE} + \lambda_{1}\mathcal{L}_{SEG}  
=\sum_{(i,j)}\left(|\hat{D}(i,j)-D(i,j)|^{2} + \lambda_1 \sum_{k=1}^c CE(M_k(i,j),\hat{M}_k(i,j)) \right), \nonumber
\label{eq:total-loss-label}
\end{align}
}}where $CE()$ denotes the cross-entropy loss, $\hat{D}$ and $\hat{M}_{k}$ are the predicted density map and segmentation respectively; $D$ and $M_{k}$ are the ground-truth density map and segmentation respectively.

For an unlabeled image, we construct an unsupervised loss based on the surrogate tasks and use it to train the feature extractor:
\begin{align}
\mathcal{L}_{U} & = \lambda_{2}\mathcal{L}_{SEG} = \lambda_{2} \sum_{k=1}^{c} \sum_{(i,j,s_{ij}) \in \mathcal{S}_k}  CE\left(\hat{M}_{k}(i,j)),s_{ij}\right),
\label{eq:total-loss-unlabel}
\end{align}
where the $\mathcal{S}_k = \{(i,j,s_{ij})\}$ denotes the set of generated pseudo labels at the $k$-th segmentation predictor. Please refer to Algorithm \ref{alg:pesudo-generate} for the generation of $\mathcal{S}_k$.

\section{Experimental Results}
We conduct extensive experiments on three popular crowd-counting datasets. The purpose is to verify if the proposed methods can achieve superior performance over other alternatives in a \textbf{semi-supervised learning setting} and understand the impact of various components of our method. Note that works that methods in a fully-supervised setting or a unsupervised setting are \textbf{not directly comparable} to ours. 
\subsection{Experimental Settings}
\noindent \textbf{Datasets.} ShanghaiTech \cite{zhang2016single}, UCF-QNRF \cite{Idrees_2018_ECCV} and WorldExpo'10 \cite{zhang2015cross} are used throughout our experiments. We modify the setting of each dataset to suit the need of semi-supervised learning evaluation. Specifically, the original training dataset is divided into labeled and unlabeled sets. The details about such partition are given as follows.

\noindent \textit{ShanghaiTech \cite{zhang2016single}}: {The ShanghaiTech dataset consists of 1,198 images with 330,165 annotated persons, which is divided into two parts: Part\_A and Part\_B. Part\_A is composed of 482 images with 244,167 annotated persons; the training set includes 300 images; the remaining 182 images are used for testing. Part\_B consists of 716 images with 88,498 annotated persons. The size of the training set is 400, and the testing set contains 316 images. In Part\_A, we randomly pick up 210 images to consist the unlabeled set, 90 images to consist the labeled set (60 images for validation). Also, In Part\_B, we randomly pick up 280 images to consist the unlabeled set, 120 images to consist the labeled set (80 images for validation).}

\noindent \textit{UCF-QNRF \cite{Idrees_2018_ECCV}}: The UCF-QNRF dataset contains 1,535 high-resolution images with 1,251,642 annotated persons. The training set includes 1,201 images, and the testing set contains 334 images. 
We randomly pick up 721 images to consist the unlabeled set, 480 images to consist the labeled set (240 images for validation).

\noindent \textit{World Expo'10 \cite{zhang2015cross}}:The World Expo'10 dataset includes 3980 frames from Shanghai 2010 WorldExpo. The training set contains 3380 images, and the testing set consists of 600 frames. Besides, the Region of Interest (ROI) is available in each scene. Each frame and the corresponding annotated person should be masked with ROI before training. We randomly pick up 2433 images to consist the unlabeled set, 947 images to consist the labeled set (271 images for validation).

\noindent \textbf{Compared Methods:} We compare the proposed IRAST method against four methods: (1) Label data only (Label-only): only use the labeled dataset to train the network. This is the baseline of all semi-supervised crowd counting approaches. (2) Learning to Rank (L2R): a semi-supervised crowd counting method proposed in \cite{liu2018leverageing}. As the unlabeled images used in this paper are not released, we re-implement it with the same backbone and test setting as our method to ensure a fair comparison. (3) Unsupervised Data Augmentation (UDA): UDA \cite{xie2019unsupervised} is one of the state-of-the-art semi-supervised learning methods. It encourages the network to generate similar predictions for an unlabeled image and its augmented version. This method was developed for image classification. We modify it by using the estimated density map as the network output. (4) Mean teacher (MT): Mean teacher \cite{tarvainen2017mean} is a classic consistency-based semi-supervised learning approach. Similar as UDA, it was originally developed for the classification task and we apply it to the regression task by changing the network work output as the estimated density maps. (5) Interpolation Consistency Training (ICT): ICT \cite{verma2019interpolation} is a recently developed semi-supervised learning approach. It is based on the mixup data augmentation \cite{zhang2017mixup} but performed on unlabeled data. Again, we tailor it for the density map regression task by changing the output as the density map. More details about the implementation of the compared methods can be found in the supplementary material.

\noindent \textbf{Implementation details:}
The feature extractor used in most of our experiment is based on the CSRNet \cite{li2018csrnet:}. We also conducted an ablation study in Section \ref{sec:use-spn} to use Scale Pyramid Network (SPN) \cite{chen2019scale} as the feature extractor. Both CSRNet and SPN leverage VGG-16 \cite{simonyan2014very} as the backbone. Also, three segmentation predictors are used by default unless specified. The thresholds for the corresponding surrogate tasks are selected as $\{0, 0.5N, 0.7N\}$ (please refer to Section \ref{sec:surrogate-task} for the method of choosing thresholds). The segmentation predictors are attached to the 14-th layer of the CSRNet or the 13-th layer of SPN. The rest layers in those networks are viewed as the task specific predictor, i.e., the density map regressor. The segmentation predictors share the same network structure as the density map regressor. Please refer the supplementary material for the detailed structure of the network.  In all experiments, we set the batch size as 1 and use Adam \cite{Kingma_2014} as the optimizer. The learning rate is initially set to 1e-6 and halves per 30 epochs (120 epochs in total). Besides, we set $t_p$ to 0.9 in experiments. Our implementation is based on PyTorch \cite{paszke2019pytorch} and we will also release the code.

\noindent \textbf{Evaluation metrics:} Following the previous works \cite{zhang2015cross,zhang2016single}, the Mean Absolute Error (MAE) and Mean Squared Error (MSE) are adopted as the metrics to evaluate the performance of the compared crowd counting methods. 

\subsection{Datasets and Results}
\subsubsection{Evaluation on the ShanghaiTech Dataset:}

The experimental results on ShanghaiTech dataset are shown in Table \ref{tab:The-performance-ShanghaiTech}. As seen, if we only use the labeled image, the network can only attain an MAE of 98.3 on Part\_A and 15.8 on Part\_B. In general, using a semi-supervised learning approach brings improvement. The L2R \cite{liu2018leverageing} shows an improvement around 8 people in the MAE of Part A but almost no improvement for Part B. Semi-supervised learning approaches modified from the classification task (UDA, MT, ICT) also lead to improved performance over Label-only on Part A. However, the improvement is not as large as L2R. Our approach, IRAST, clearly demonstrates the best performance. It leads to 11.4 MAE improvement over the Label-only on Part A.

\begin{table*}\scriptsize
\begin{floatrow}
\capbtabbox{
\begin{tabular}{cccccc}
\hline
 & \multirow{2}{*}{Method} &  \multicolumn{2}{l}{Part\_A} & \multicolumn{2}{l}{Part\_B} \\ \cline{3-6} 
 &  & MAE & MSE & MAE & MSE \\ \hline

\multirow{5}{*}{\rotatebox[origin=c]{90}{Semi}} 

 & Label-only & 98.3 & 159.2 & 15.8 & 25.0  \\ \cline{2-6}
  & L2R \cite{liu2018leverageing}& 90.3 & 153.5 & 15.6 & 24.4 \\ \cline{2-6} 
& UDA \cite{xie2019unsupervised} & 93.8 & 157.2 & 15.7 & 24.1 \\ \cline{2-6} 
 & MT \cite{tarvainen2017mean}& 94.5 & 156.1 & 15.6 & 24.5 \\ \cline{2-6} 
 & ICT \cite{verma2019interpolation}& 92.5 & 156.8 & 15.4 & 23.8 \\ \cline{2-6} 
 & IRAST & \textbf{86.9} & \textbf{148.9} & \textbf{14.7} & \textbf{22.9} \\ \hline
 \multirow{1}{*}{} 
 & (Fully) CSRNet \cite{li2018csrnet:}& 68.2 & 115.0 & 10.6 & 16.0  \\ \hline

\end{tabular}
}{
\caption{The comparison on the ShanghaiTech dataset. The best results are in bold font.}
\label{tab:The-performance-ShanghaiTech}
}
\capbtabbox{
\begin{tabular}{cccc}
\hline
 &\multirow{2}{*}{Method} & \multicolumn{2}{l}{UCF-QNRF} \\ \cline{3-4}
\multicolumn{1}{l}{} &  & MAE & MSE \\ \hline

\multirow{5}{*}{\rotatebox[origin=c]{90}{Semi}} 
 & Label-only & 147.7 & 253.1 \\ \cline{2-4}
 & L2R  \cite{liu2018leverageing}& 148.9 & 249.8 \\ \cline{2-4} 
 & UDA \cite{xie2019unsupervised} & 144.7 & 255.9 \\ \cline{2-4} 
 & MT \cite{tarvainen2017mean}& 145.5 & 250.3 \\ \cline{2-4} 
 & ICT \cite{verma2019interpolation} & 144.9 & 250.0 \\ \cline{2-4} 
 & IRAST & \textbf{135.6} & \textbf{233.4} \\ \hline
  \multirow{1}{*}{} 
 & (Fully) CSRNet \cite{li2018csrnet:}& 119.2 & 211.4 \\ \hline
\end{tabular}
}{
\caption{The comparison on the UCF-QNRF dataset. The best results are in bold font.}
\label{tab:The-performance-UCF-QNRF}
}
\end{floatrow}
\end{table*}

\subsubsection{Evaluation on the UCF-QNRF Dataset:} 

% \vspace{-40pt}
\begin{figure*}[tbh]
\includegraphics[width=12.6cm]{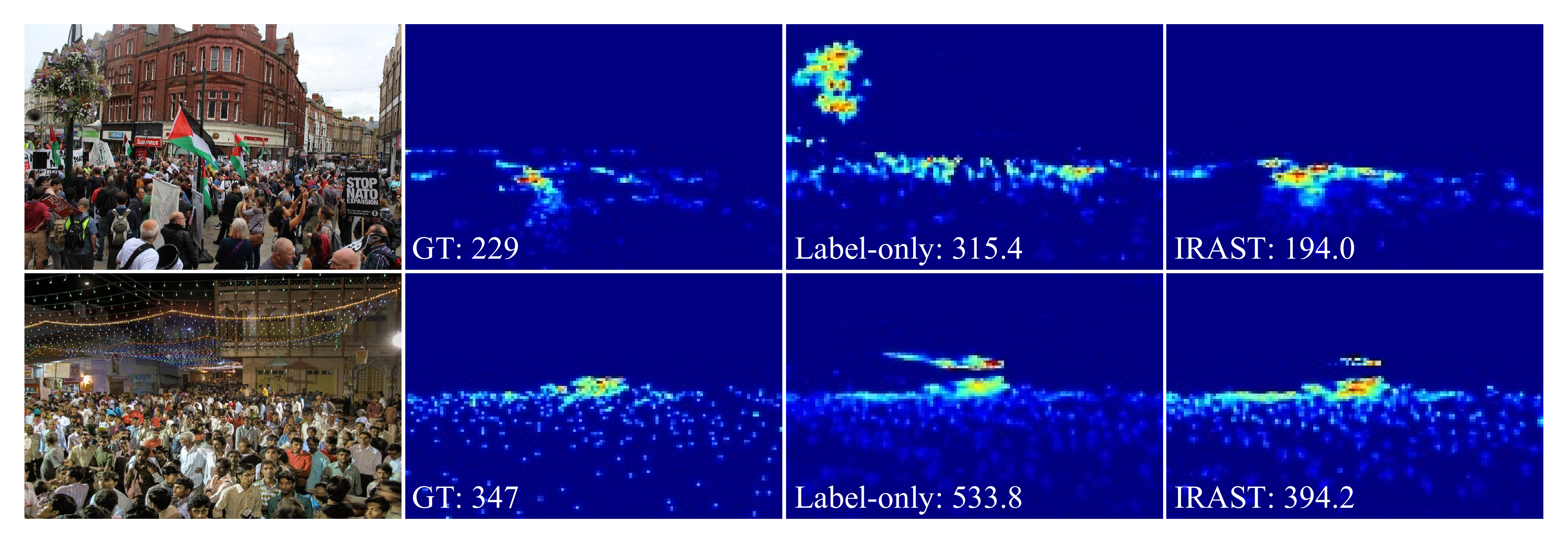}\caption{A comparison of predicted density maps on the UCF-QNRF dataset.     \label{fig:visial_ucf}}
\end{figure*}

The advantage of the proposed method is also well demonstrated on UCF-QNRF dataset, shown in Table \ref{tab:The-performance-UCF-QNRF}. Again, the proposed method achieves the overall best performance, and exceeds the Label-only by around 12 MAE. The other semi-supervised learning approach does not work well on this dataset. In particular, L2R even achieves worse performance than the Label-only. This on the other hand clearly demonstrates the robustness of our approach. Also, from the results in both ShanghaiTech and UCF-QNRF, we can see that directly employing the semi-supervised learning approaches which were originally developed for classification may not achieve satisfying performance. It remains challenging for developing the semi-supervised crowd counting algorithm.

\begin{table}\scriptsize
\centering{}\caption{The performance comparison in terms of MAE on the WorldExpo'10 dataset. The best results are in bold font. \label{tab:The-performance-comparison-WorldExpo}}

\begin{tabular}{cccccccc}
\hline 
\multirow{1}{*}{} & {Method} & Sce.1 & {Sce.2} & {Sce.3} & {Sce.4} & {Sce.5} & {Avg.}\tabularnewline
\hline 

\multirow{6}{*}{{\rotatebox[origin=c]{90}{Semi}}}

&{Label-only} & {2.4} & {16.9} & {9.7} & {41.3} & {3.1} & {14.7}\tabularnewline
\cline{2-8} \cline{3-8} \cline{4-8} \cline{5-8} \cline{6-8} \cline{7-8} \cline{8-8} 
 & {L2R \cite{liu2018leverageing}} & {2.4} & {20.9} & {9.8} & {31.9} & {4.4} & {13.9}\tabularnewline
 \cline{2-8} \cline{3-8} \cline{4-8} \cline{5-8} \cline{6-8} \cline{7-8} \cline{8-8} 
& {UDA \cite{xie2019unsupervised}} & \textbf{1.9} & {20.3} & {10.9} & {34.5} & {3.6} & {14.2}\tabularnewline
 \cline{2-8} \cline{3-8} \cline{4-8} \cline{5-8} \cline{6-8} \cline{7-8} \cline{8-8} 
 & {MT \cite{tarvainen2017mean}} & {2.6} & {24.8} & {9.4} & {30.3} & {3.3} & {14.1}\tabularnewline
 \cline{2-8} \cline{3-8} \cline{4-8} \cline{5-8} \cline{6-8} \cline{7-8} \cline{8-8} 
 & {ICT \cite{verma2019interpolation}} & {2.3} & {17.8} & \textbf{{8.3}} & {43.5} & \textbf{2.8} & {14.9}\tabularnewline
  \cline{2-8} \cline{3-8} \cline{4-8} \cline{5-8} \cline{6-8} \cline{7-8} \cline{8-8} 
 & {IRAST} & {2.2} & \textbf{12.3} & {9.2} & \textbf{{27.8}} & {4.1} & \textbf{{11.1}}\tabularnewline
\hline 
\multirow{1}{*}{}
& (Fully) {CSRNet} \cite{li2018csrnet:} & {2.9} & {11.5} & {8.6} & {16.6} & {3.4} &
{8.6}\tabularnewline
\hline 
\end{tabular}
\end{table}
\subsubsection{Evaluation on the World Expo'10 Dataset:}
 
The results are shown in Table \ref{tab:The-performance-comparison-WorldExpo}. As seen, IRAST again achieves the best MAE in 2 scenes and delivers the best MAE over other methods. The other semi-supervised learning methods achieve comparable performance and their performance gain over the Label-only is not significant. 

\subsection{Ablation Study}
To understand the importance of various components in our algorithm, we conduct a serials of ablation studies. 

\subsubsection{Varying the Number of Labeled Images:}

% \vspace{-30pt}
\begin{figure}
\includegraphics[width=6cm]{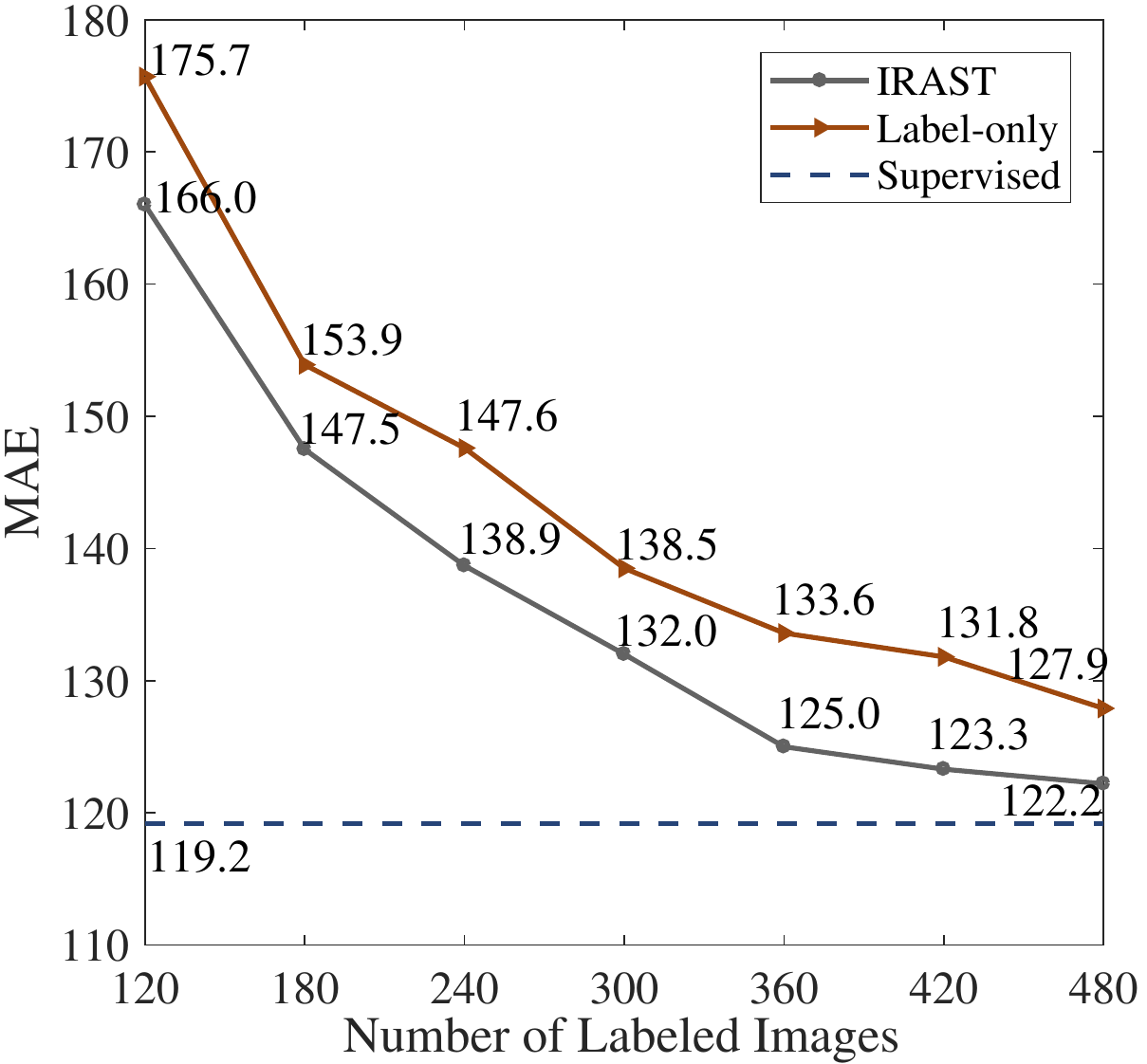}
\caption{The impact of the number of labeled images. Evaluated in terms of MAE on the UCF-QNRF dataset. \label{fig:alter-label-num}}
\end{figure}

We first examine the performance gain over the Label-only under different amount of labeled images. We conduct experiments on the UCF-QNRF dataset. We vary the number of labeled image from 120 to 480 while fixing the amount of unlabeled images to be 481. The performance curves of the Label-only and IRAST are depicted in Figure \ref{fig:alter-label-num}. As seen, IRAST achieves consistent performance gain over the Label-only, which is an evidence of the robustness of our method. Also, we can see that with IRAST, using 480 images can almost achieve comparable performance than the performance of a fully-supervised model which needs 961 training images.

\subsubsection{IRAST on Labeled set:}
The proposed method constructs an additional training task and one may suspect the good performance is benefited from the multi-task learning. To investigate this hypothesis, we also conduct an ablation study by learning the crowd counter on the labeled set only, but with both density map regression task and surrogate tasks. The results are shown in Table \ref{tab:Results-of-only-label}. As seen, using multiple-surrogate tasks for the labeled set does improve the performance to some extent, but still has a significant performance gap with the proposed method. This result clearly validates that our method can not be simply understood as a multi-task learning approach. 

\subsubsection{Other Alternative Surrogate Task:\label{sec:msst-vs-irast}}
One alternative method is to use a multi-class segmentation predictor to train the feature extractor, namely MSST mentioned in Section \ref{sec:inner-aware}. To compare MSST and IRAST, we conduct experiments on the ShanghaiTech Part\_A and UCF-QNRF dataset. The results are shown in Table \ref{tab:Results-compare-msst-IRAST}. As seen, MSST can achieve a better performance than Label-only method, which demonstrates the effectiveness of using a surrogate task for feature learning. However, MSST obtains a worse crowd counting performance than IRAST. Recall that MSST lacks an error correction mechanism to generate pseud-label, the superior performance of IRAST over MSST provides evidence to support the merit of our multiple surrogate binary-segmentation task modelling. 

\vspace{-15pt}
\begin{table*}\scriptsize
\begin{floatrow}
\capbtabbox{
\begin{tabular}{ccccc}
\hline
\multirow{2}{*}{Method} & \multicolumn{2}{l}{Part\_A} & \multicolumn{2}{l}{UCF-QNRF} \\ \cline{2-5} 
 & MAE & MSE & MAE & MSE \\ \hline
Label-only & 98.3 & 159.2 & 147.7 & 253.1 \\ \hline
IRAST on label & 94.1 & 151.6 & 140.8 & 245.4 \\ \hline
IRAST & \textbf{86.9} & \textbf{148.9} & \textbf{135.6} & \textbf{233.4} \\ \hline
\end{tabular}
}{
\caption{Impact of the unlabeled images in the process of feature learning. Evaluated on the ShanghaiTech Part\_A and UCF-QNRF dataset. The best results are in bold font.}
\label{tab:Results-of-only-label}
}
\capbtabbox{
\begin{tabular}{ccccc}
\hline
\multirow{2}{*}{Method} & \multicolumn{2}{l}{Part\_A} & \multicolumn{2}{l}{UCF-QNRF} \\ \cline{2-5} 
 & MAE & MSE & MAE & MSE \\ \hline
Label-only & 98.3 & 159.2 & 147.7 & 253.1 \\ \hline
MSST & 91.5 & 155.2 & 140.0 & 233.7 \\ \hline
IRAST & \textbf{86.9} & \textbf{148.9} & \textbf{135.6} & \textbf{233.4} \\ \hline
\end{tabular}
}{
\caption{Comparison of IRAST and MSST. Evaluated on the ShanghaiTech Part\_A and UCF-QNRF dataset. The best results are in bold font.}
\label{tab:Results-compare-msst-IRAST}
}
\end{floatrow}
\end{table*}

\vspace{-30pt} 
\begin{table*}\scriptsize
\begin{floatrow}
\capbtabbox{
\begin{tabular}{ccccc}
\hline
\multirow{2}{*}{Method} & \multicolumn{2}{l}{Part\_A} & \multicolumn{2}{l}{UCF-QNRF} \\ \cline{2-5} 
 & MAE & MSE & MAE & MSE \\ \hline
Label-only & 98.3 & 159.2 & 147.7 & 253.1 \\ \hline
IRAST w/o IR & 93.5 & 155.5 & 139.8 & 240.3 \\ \hline
IRAST & \textbf{86.9} & \textbf{148.9} & \textbf{135.6} & \textbf{233.4} \\ \hline
\end{tabular}}{
\caption{Impact of the inter-relationship. Evaluated on the ShanghaiTech Part\_A and UCF-QNRF dataset. The best results are in bold font.\label{tab:Results-wo-inter-relationship}}
}

\capbtabbox{
\begin{tabular}{ccccc}
\hline
\multirow{2}{*}{Method} & \multicolumn{2}{l}{Part\_A} & \multicolumn{2}{l}{UCF-QNRF} \\ \cline{2-5} 
 & MAE & MSE & MAE & MSE \\ \hline
Label-only & 98.3 & 159.2 & 147.7 & 253.1 \\ \hline
$t_p=0.6$ & 88.4 & 152.3 & 137.2 & 234.9 \\ \hline
$t_p=0.9$ & \textbf{86.9} & \textbf{148.9} & \textbf{135.6} & \textbf{233.4} \\ \hline
\end{tabular}
}{
\caption{Impact of the changing hyper-parameter $t_p$. Evaluated on the ShanghaiTech Part\_A and UCF-QNRF dataset. The best results are in bold font. \label{tab:change-tp}}}
\end{floatrow}
\end{table*}

\subsubsection{The Importance of Considering the Inter-Relationship:}
In IRAST, we leverage the Inter-Relationship (IR) between surrogate tasks to generate pseudo-labels. To verify the importance of this consideration, we conduct an ablation study by removing the inter-Relationship constraint for pseudo-label generation. The results are shown in Table \ref{tab:Results-wo-inter-relationship}. As seen, a decrease in performance is observed when the Inter-Relationship is not considered. This observation suggests that the Inter-Relationship awareness is essential to the proposed IRAST method.

\subsubsection{The Impact of Changing the Prediction  Confidence  Threshold:}
We set hyper-parameter $t_p$ to 0.9 in the previous experiments. To investigate the impact of $t_p$, we conduct experiments on ShanghaiTech part\_A and UCF-QNRF dataset. The results are shown in Table \ref{tab:change-tp}. The results demonstrate setting diverse $t_p$ does not impact crowd counting performance significantly. The crowd counting performance are comparable to our current results. It means the proposed IRAST method is robust.

\vspace{-10pt} 
\begin{table*}\scriptsize
\begin{floatrow}
\capbtabbox{
\begin{tabular}{ccccc}
\hline
\multirow{2}{*}{Method} & \multicolumn{2}{c}{Part\_A} & \multicolumn{2}{c}{UCF-QNRF} \\ \cline{2-5} 
 & MAE & MSE & MAE & MSE \\ \hline
Label-only (CSRNet) & 98.3 & 159.2 & 147.7 & 253.1 \\ \hline
IRAST (CSRNet) & 86.9 & 148.9 & 135.6 & 233.4 \\ \hline
Label-only (SPN) & 88.5 & 152.6 & 138.0 & 244.5 \\ \hline
IRAST (SPN) & \textbf{83.9} & \textbf{140.1} & \textbf{128.4} & \textbf{225.3} \\ \hline
\end{tabular}}{
\caption{Impact of the feature extractor. Evaluated on the ShanghaiTech Part\_A and UCF-QNRF dataset. The best results are in bold font.\label{tab:change-backbone}}
}
\capbtabbox{
\begin{tabular}{ccccc}
\hline
\multirow{2}{*}{Tasks} & \multicolumn{2}{c}{Part\_A} & \multicolumn{2}{c}{UCF-QNRF} \\ \cline{2-5} 
 & MAE & MSE & MAE & MSE \\ \hline
1 & 89.8 & 149.8 & 142.8 & 236.5 \\ \hline
2 & 88.9 & 149.6 & 139.1 & 237.8 \\ \hline
\textbf{3} & \textbf{86.9} & \textbf{148.9} & \textbf{135.6} & \textbf{233.4} \\ \hline
4 & 90.1 & 150.2 & 137.5 & 236.8 \\ \hline
5 & 90.3 & 150.9 & 137.8 & 234.4\\ \hline
\end{tabular}
}{
\caption{Impact of the varing number of surrogate tasks. The best results are in bold font. \label{tab:change-surrogate-tasks-num}}}
\end{floatrow}
\end{table*}

\subsubsection{Change of the Feature Extractor:\label{sec:use-spn}}
So far, we conduct our experiment with the CSRNet \cite{li2018csrnet:} feature extractor. It is unclear if performance gain can still be achieved with other feature extractors. To investigate this, we conduct an experiment that uses SPN \cite{chen2019scale} as the feature extractor on the ShanghaiTech Part\_A and UCF-QNRF dataset. Results are shown in Table \ref{tab:change-backbone}. We can see that the significant performance gain can still be achieved. Also, we observe an improved overall performance by using SPN. This suggests that the advances in network architecture design for crowd counting can be readily incorporated into our method.

\subsubsection{The Effect of Varying the Number of Surrogates Tasks:}
Finally, we test the impact of choosing the number of surrogate tasks. We incrementally adding more thresholds by following the threshold sequence $\{0,0.5N,0.7N,0.8N,0.9N\}$, e.g., $\{0,0.5N,0.7N\}$ is used for the three-task setting while $\{0,0.5N,0.7N,0.8N\}$ is used for the four-task setting. The results are shown in Table \ref{tab:change-surrogate-tasks-num}. The results demonstrate setting three surrogate tasks for feature learning can achieve the best crowd counting performance. To have a finer grained partition of density value does not necessarily lead to improved performance. 

\section{Conclusions}
In this paper, we proposed a semi-supervised crowd counting algorithm by creating a set of surrogate tasks for learning the feature extractor. A novel self-training strategy that can leverage the inter-relationship of different surrogate tasks is developed. Through extensive experiments, it is clear that the proposed method enjoys superior performance over other semi-supervised crowd counter learning approaches. 

\section*{Acknowledgement} 
This work was supported by the Key Research and Development Program of Sichuan Province (2019YFG0409). Lingqiao Liu was in part supported by ARC DECRA Fellowship DE170101259.

\bibliographystyle{unsrt}
\bibliography{egbib}

\clearpage 

\section*{Supplementary}
This file provides some additional information from two perspective: compared semi-supervised methods and structures of used crowd counting networks,
which correspond to the Section 6 in the paper.

\subsection*{Introduction of Compared Methods}
\subsubsection*{Learning to Rank (L2R):}
is proposed in \cite{liu2018leverageing}. This method collects a large amount of unlabeled extra images from the Internet for density estimation and produces a serial of sub-images $I_1,I_2,...,I_M$ and ensures the smaller sub-image $I_t$ is covered in the larger sub-image $I_s$. The number of pedestrians in the smaller sub-image is not greater than the larger one, though the exact person counts is unknown. It can be used for crowd counting model training as shown in Eq. \ref{eq:rank-loss}.
\begin{equation}
\mathcal{L}_{r}=\sum_{s=1}^{M}\sum_{t=1}^{s} max(0,\hat{C_t}-\hat{C_s}),\label{eq:rank-loss}
\end{equation}
where $\hat{C_t}$ and $\hat{C_s}$ are the estimated count value in the $t-th$ and $s-th$ sub-image. Note that the $t-th$ sub-image is contained in the $s-th$ sub-image.

\subsubsection*{Unsupervised Data Augmentation (UDA):} is a classical consistency-based   semi-supervised learning method for classification. The core of this method is to create an augmented version of the original unlabeled image and reduce the diversity of predicted density maps directly. It can be formulated as:

\begin{equation}
\mathcal{L}_{UDA}=\sum_{(i,j)}|\hat{D}_{A}(i,j)-\hat{D}(i,j)|^{2},\label{eq:UDA-loss}
\end{equation}
where $\hat{D}_{A}$ is the estimated density map from augmented version of unlabeled images and $\hat{D}$ is the predicted density map from original unlabeled images.

\subsubsection*{Mean Teacher (MT):} is widely used in semi-supervised learning tasks. This method defines a teacher model and student model, these two models have the same structure. For unlabeled images, the student model uses the original image to produce a predicted density map while the teacher model leverage an augmented version of the unlabeled image to produce another version of the predicted density map. The process of updating the parameters of the student model can be formulated as:

\begin{equation}
\mathcal{L}_{MT_S}=\sum_{(i,j)}|\hat{D}_{S}(i,j)-\hat{D}_{T}(i,j)|^{2},\label{eq:MT-S}
\end{equation}
where $\hat{D}_{S}$ is the predicted density map from the student model, $\hat{D}_{T}$ is the predicted density map from the teacher model. Besides, unlike the student model, the update process of teacher model averages model weights which is formulated as:
\begin{equation}
\mathcal{\theta}_{t}^{T}=\alpha{\theta}_{t-1}^{T}+(1-\alpha){\theta}_{t}^{S},\label{eq:MT-T}
\end{equation}
where ${\theta}_{t}^{T}$ is the parameter of the teacher model at the $t-th$ steps, $\alpha$ is a smoothing coefficient hyper parameter, ${\theta}_{t-1}^{T}$ is the parameter of the teacher model at the $(t-1)-th$ steps and ${\theta}_{t}^{S}$ is the parameter of the student model currently.

\subsubsection*{Interpolation Consistency Training (ICT):} is a semi-supervised learning method for classification. The core of this method is using the original unlabeled image and the augmented version to train the crowd counting model. The training loss is formulated as:

\begin{equation}
\mathcal{L}_{ICT}=\sum_{(i,j)}|\hat{D}_{\lambda}(i,j)-\lambda(\hat{D}_{A}(i,j),\hat{D}(i,j)|^{2},\label{eq:ICT}
\end{equation}
where $\lambda$ is the mix-up function. $D_\lambda$ is the predicted density map produced by the mix-up version of the unlabeled image which is $I_\lambda= \lambda I_A + (1-\lambda) I$. The $\lambda(\hat{D}_{A}(i,j),\hat{D}(i,j)$ is the mix-up version of predicted density maps which is equal to $\lambda D_A + (1-\lambda) D$.

\subsection*{Introduction of Network structures}
In the paper, the crowd counting model includes the feature extractor, the density regressor and the segmentation predictor. In our experiment, we use the CSRNet \cite{li2018csrnet:} and the SPN \cite{chen2019scale}. The details of these structures are shown in Table. \ref{tab:structure}. The segmentation predictor has a similar structure with the corresponding density regressor expect the output of the last layer contains two channels. 

\begin{table}[]
\centering{}\caption{The structure of the CSRNet and the SPN. The k is the convolutional kernel size, c means the number of output channel, s means stride, p means padding size and d indicates the dilation rate. \label{tab:structure}}
\begin{tabular}{ccc}
\hline & CSRNET  & \multicolumn{1}{c|}{SPN}                                         \\ \hline
\multirow{17}{*}{\rotatebox{90}{Feature Extractor}} & k(3,3)-c64-s1-p1-d1  & k(3,3)-c64-s1-p1-d1 \\
& k(3,3)-c64-s1-p1-d1  & k(3,3)-c64-s1-p1-d1\\
 & maxpooling(2,2)  & maxpooling(2,2)  \\& k(3,3)-c128-s1-p1-d1 & k(3,3)-c128-s1-p1-d1 \\& k(3,3)-c128-s1-p1-d1 & k(3,3)-c128-s1-p1-d1 \\
 & maxpooling(2,2)   & maxpooling(2,2) \\ & k(3,3)-c256-s1-p1-d1 & k(3,3)-c256-s1-p1-d1\\ & k(3,3)-c256-s1-p1-d1 & k(3,3)-c256-s1-p1-d1 \\ & k(3,3)-c256-s1-p1-d1 & k(3,3)-c256-s1-p1-d1\\ & maxpooling(2,2)  & maxpooling(2,2) \\ & k(3,3)-c512-s1-p1-d1 & k(3,3)-c512-s1-p1-d1  \\& k(3,3)-c512-s1-p1-d1 & k(3,3)-c512-s1-p1-d1 \\  & k(3,3)-c512-s1-p1-d1 & k(3,3)-c512-s1-p1-d1\\ & k(3,3)-c512-s1-p1-d2 & \begin{tabular}[c]{@{}c@{}}{[}k(3,3)-c512-s1-p2-d2,\\ k(3,3)-c512-s1-p4-d4,\\ k(3,3)-c512-s1-p8-d8,\\ k(3,3)-c512-s1-p12-d12{]}\end{tabular} \\
 & k(3,3)-c512-s1-p2-d2 & k(3,3)-c512-s1-p1-d1  \\  & k(3,3)-c512-s1-p2-d2 & k(3,3)-c256-s1-p1-d1\\   & k(3,3)-c256-s1-p2-d2 &      \\ \hline
\multirow{3}{*}{Density regressor} & k(3,3)-c128-s1-p2-d2 & k(3,3)-c128-s1-p1-d1 \\ & k(3,3)-c64-s1-p2-d2  & k(3,3)-c64-s1-p1-d1 \\  & k(1,1)-c1-s1-p0-d0   & k(1,1)-c1-s1-p0-d0      \\\hline
\end{tabular}
\end{table}

\end{document}